%% file: main.tex
\def\year{2021}\relax
\documentclass[letterpaper]{article} 
\usepackage{aaai21}  
\usepackage{times}  
\usepackage{helvet} 
\usepackage{courier}  
\usepackage[hyphens]{url}  
\usepackage{graphicx} 
\usepackage{import}

\usepackage[switch]{lineno}
\usepackage{natbib}  
\usepackage{caption} 
\usepackage{times}
\usepackage{epsfig}
\usepackage{graphicx}
\usepackage{amsmath}
\usepackage{bbm}
\usepackage{amssymb}
\usepackage{amssymb}
\usepackage{multirow}
\usepackage{mathtools}
\usepackage{subcaption}
\usepackage{float}
\usepackage{todonotes}

\usepackage{bm}
\usepackage{pifont}
\newcommand{\cmark}{\text{\ding{51}}}

\newcommand{\etal}{\emph{et~al}}

\title{A Hybrid Attention Mechanism for Weakly-Supervised Temporal Action Localization}
\author {
    Ashraful Islam\textsuperscript{\rm 1},
    Chengjiang Long \textsuperscript{\rm 2},
    Richard Radke \textsuperscript{\rm 1} \\
}
\affiliations {
    \textsuperscript{\rm 1} Rensselaer Polytechnic Institute \\
    \textsuperscript{\rm 2} JD Digits AI Lab \\
    islama6@rpi.edu,  chengjiang.long@jd.com, rjradke@ecse.rpi.edu
}

\begin{document}

\maketitle

\begin{abstract}
Weakly supervised temporal action localization is a challenging vision task due to the absence of ground-truth temporal locations of actions in the training videos. With only video-level supervision during training, most existing methods rely on a Multiple Instance Learning (MIL) framework to predict the start and end frame of each action category in a video. However, the existing MIL-based approach has a major limitation of only capturing the most discriminative frames of an action, ignoring the full extent of an activity. Moreover, these methods  cannot model background activity effectively, which plays an important role in localizing foreground activities. In this paper, we present a novel framework named HAM-Net with a hybrid attention mechanism which includes temporal soft, semi-soft and hard attentions to address these issues. Our temporal soft attention module, guided by an auxiliary background class in the classification module, models the background activity by introducing an ``action-ness'' score for each video snippet. Moreover, our temporal semi-soft and hard attention modules, calculating two attention scores for each video snippet, help to focus on the less discriminative frames of an action to capture the full action boundary. Our proposed approach outperforms recent state-of-the-art methods by at least 2.2\% mAP at IoU threshold 0.5 on the THUMOS14 dataset, and by at least 1.3\% mAP at IoU threshold 0.75 on the ActivityNet1.2 dataset. Code can be found at: \url{https://github.com/asrafulashiq/hamnet}. 

\end{abstract}

\section{Introduction} \label{sec:intro}

Temporal action localization refers to the task of predicting the start and end times of all action instances in a video. There has been remarkable progress in fully-supervised temporal action localization \cite{Tran2018ACL, ssn, frcnn, lin2018bsn, xu2019gtad}. However, annotating the precise temporal ranges of all action instances in a video dataset is expensive, time-consuming, and error-prone. On the contrary, weakly supervised temporal action localization (WTAL) can greatly simplify the data collection and annotation cost.

\begin{figure}[!ht]
    \centering
    \includegraphics[width=0.47\textwidth]{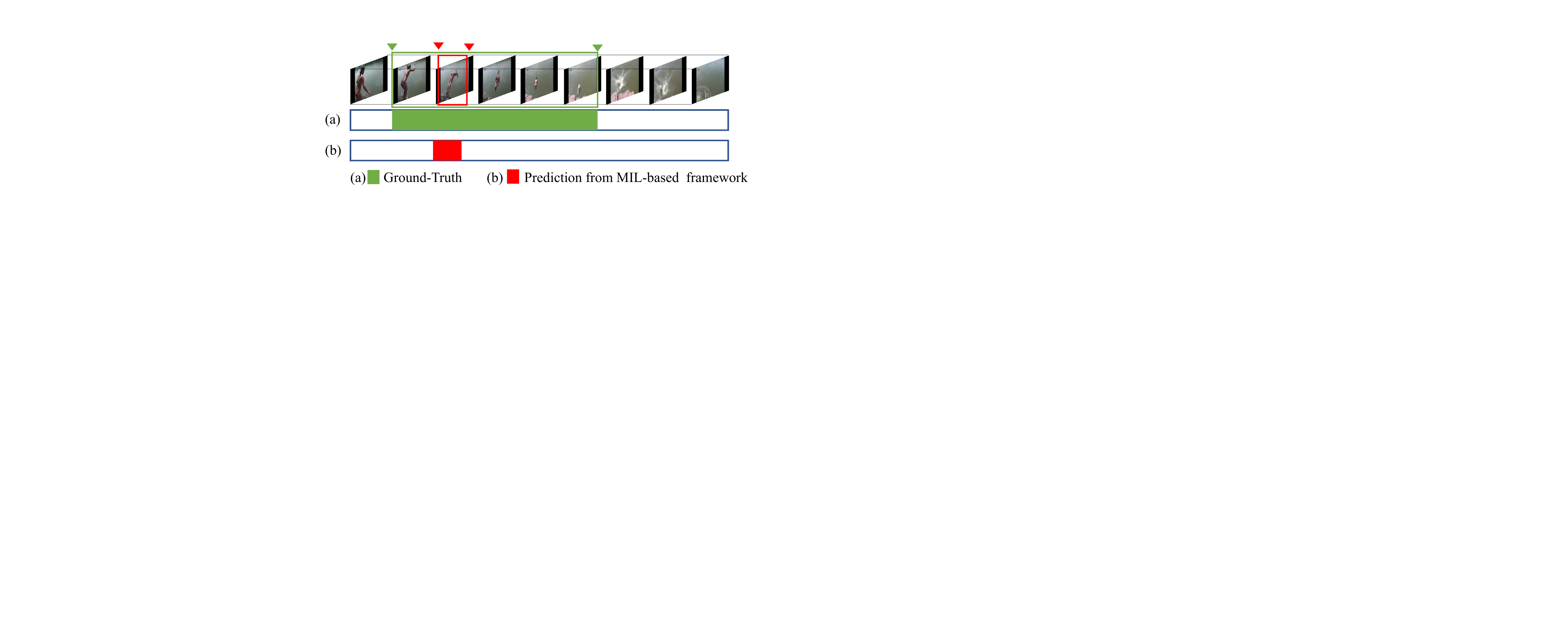}
    \caption{The existing MIL framework does not necessarily capture the full extent of an action instance. In this example of a diving activity. (a) shows the ground-truth localization, and (b) shows the prediction from an MIL-based WTAL framework. The MIL framework only captures the most discriminative part of the diving activity, ignoring the beginning and ending parts of the full action.}
    \label{fig:intro}
    \vspace{-0.4cm}
\end{figure}

WTAL aims at localizing and classifying all action instances in a video given only video-level category label during training stage. Most existing WTAL methods rely on the multiple instance learning (MIL) paradigm \cite{wtalc, liu2019completeness, islam2020weakly}. In this paradigm, a video consists of several snippets; snippet-level class scores, commonly known as Class Activation Sequences (CAS),  are calculated and then temporally pooled to obtain video-level class scores. The action proposals are generated by thresholding the snippet-level class scores.  However, this framework has a major issue: it does not necessarily capture the full extent of an action instance. As training is performed to minimize the video-level classification loss, the network predicts higher CAS values for the discriminative parts of actions, ignoring the less discriminative parts. For example, an action might consist of several sub-actions \cite{hou2017real}. In the MIL paradigm, only a particular sub-action might be detected, ignoring the other parts of the action. 

An illustrative example of a diving activity is presented in Figure \ref{fig:intro}. We observe that only the most discriminative location of the full diving activity is captured by the MIL framework. Capturing only the most distinctive part of an action is sufficient to produce a high video-level classification accuracy, but does not necessarily result in good temporal localization performance. 
Another issue with the existing framework is modeling the background activity effectively so that background frames are not included in the temporal localization prediction. It has been shown previously that background activity plays an important role in action localization \cite{lee2020backgroundbasnet}.
Without differentiating the background frames from the foreground ones, the network might include the background frames to minimize foreground classification loss, resulting in many false positive localization predictions.  

In this paper, we propose a new WTAL framework named HAM-Net with a hybrid attention mechanism to solve the above-mentioned issues. Attention mechanism has been used in successfully used in deep learning \cite{Islam_2020_DOAGAN, vaswani2017attention, Shi_2020_CVPR_attention_modeling}. HAM-Net produces soft, semi-soft and hard attentions to detect the full temporal span of action instances and to model background activity, as illustrated in Figure~\ref{fig:full_model}.

Our framework consists of (1) a classification branch that predicts class activation scores for all action instances including background activity, and (2) an attention branch that predicts the ``action-ness'' scores of a video snippet. The snippet-level class activation scores are also modulated by three snippet-level attention scores, and temporally pooled to produce video-level class scores.

To capture the full action instance, we drop the more discriminative parts of the video, and focus on the less discriminative parts. We achieve this by calculating semi-soft attention scores and hard attention scores for all snippets in the video. The semi-soft attention scores drop the more discriminative portions of the video by assigning zero value to the snippets that have soft-attention score greater than a threshold, and the scores for the other portions remain the same as the soft-attention scores. 
The video-level classification scores guided by the semi-soft attentions contain only foreground classes. On the other hand, the hard-attention score drops the more discriminative parts of the video, and assigns the attention scores of the less discriminative parts to one, which ensures that video-level class scores guided by this hard attention contain both foreground and background classes. Both the semi-soft and hard attentions encourage the model to learn the full temporal boundary of an action in the video.

To summarize, our contributions are threefold: (1) we propose a novel framework with a hybrid attention mechanism to model an action in its entirety; (2) we present a background modeling strategy by attention scores guided using an auxiliary background class; and (3) we achieve state-of-the-art performance on both the THUMOS14 \cite{THUMOS14} and ActivityNet \cite{activitynet} datasets. Specifically, we outperform  state-of-the-art methods by 2.2\% mAP at IoU threshold 0.5 on the THUMOS14 dataset, and 1.3\% mAP at IoU threshold 0.75 on the ActivityNet1.2 dataset.

\section{Related Work} \label{sec:related_work}


\subsubsection{Action Analysis with Full Supervision} Due to the representation capability of deep learning based models, and the availability of large scale datasets  \cite{THUMOS14, activitynet, charades, AVA, kinetics}, significant progress has been made in the domain of video action recognition. To design motion cues, the two-stream network \cite{simonyan2014two} incorporated optical flow \cite{horn1981opticalflow} as a separate stream along with RGB frames. 3D convolutional networks have demonstrated better representations for video \cite{i3d, c3d, Tran2018ACL}. For fully-supervised temporal action localization, several recent methods adopt a two-stage strategy  \cite{Tran2018ACL, ssn, frcnn, lin2018bsn}.

\subsubsection{Weakly Supervised Temporal Action Localization}
In terms of exisiting WTAL methods, UntrimmedNets \cite{wang2017untrimmednets} introduced a classification module for predicting a classification score for each snippet, and a selection module to select relevant video segments. On top of that, STPN \cite{stpn} added a sparsity loss and class-specific proposals. AutoLoc \cite{shou2018autoloc} introduced the outer-inner contrastive loss to effectively predict the temporal boundaries. W-TALC \cite{wtalc} and Islam and Radke  \cite{islam2020weakly}  incorporated distance metric learning strategies.

MAAN \cite{yuan2018maan} proposed a novel marginalized average aggregation module and latent discriminative probabilities to reduce the difference between the most salient regions and the others. TSM \cite{yu2019weaktsm} modeled each action instance as a multi-phase process  to effectively characterize action instances. WSGN \cite{fernando2020weaklywsgn} assigns a weight to each frame prediction based on both local and global statistics. DGAM \cite{Shi_2020_CVPR_attention_modeling} used a conditional Variational Auto-Encoder (VAE) to separate the attention, action, and non-action frames. CleanNet \cite{liu2019weaklycleannet} introduced an action proposal evaluator that provides pseudo-supervision by leveraging the temporal contrast in snippets. 3C-Net \cite{narayan20193cnet} adopted three loss terms to ensure the separability, to enhance discriminability, and to delineate adjacent action sequences. Moreover, BaS-Net \cite{lee2020backgroundbasnet} and Nguyen \etal  \cite{nguyen2019weaklybackgroundmodeling} modeled background activity by introducing an auxiliary background class. However, none of these approaches explicitly resolve the issue of modeling an action instance in its entirety.

To model action completeness, Hide-and-Seek \cite{Singh2017HideandSeekFA} hid part of the video to discover other relevant parts, and Liu \etal  \cite{liu2019completeness} proposed a multi-branch network where each branch predicts distinctive action parts. Our approach has similar motivation, but differs in that we hide the most discriminative parts of the video instead of random parts.

\section{Proposed Method} \label{sec:proposed_method}

\begin{figure*}[!ht]
    \centering
    \includegraphics[width=1\textwidth]{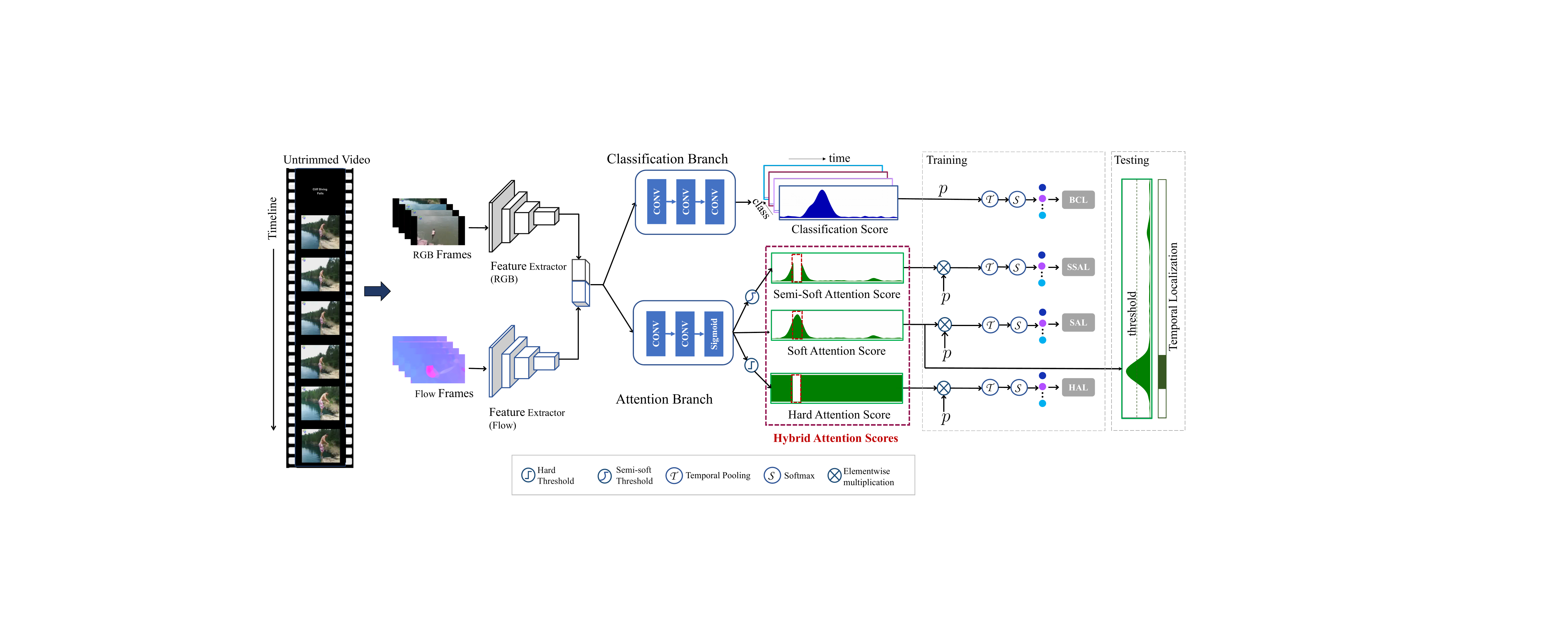}
    \caption{Overview of our proposed framework HAM-Net. Snippet-level features of both RGB and flow frames are extracted and separately fed into a classification branch and an attention branch with a hybrid attention mechanism. Three attention scores are calculated: soft attention, semi-soft attention, and hard attention, which are multiplied with snippet-level classification scores to obtain attention-guided class scores. The network is trained using four attention-guided losses: base classification loss (BCL), soft attention loss (SAL), semi-soft attention loss (SSAL), and hard attention loss (HAL), as well as sparsity loss and guide loss. }
    \label{fig:full_model}
\end{figure*}

\subsection{Problem Formulation}
Assume a training video $V$  containing activity instances chosen from $n_c$ activity classes. A particular activity can occur in the video multiple times. 
Only the video-level action instances are given. Denote the video-level activity instances as $\mathbf{y} \in \{0,1\}^{n_c}$, where $y_{j}=1$ only if there is at least one instance of the $j$-th action class in the video, and $y_{j}=0$ if there is no instance of the $j$-th activity. Note that neither the frequency nor the order of the action instances in the video is provided. Our goal is to create a model that is trained only with video-level action classes, and predicts temporal location of activity instances during evaluation, {\em i.e.}, for a testing video it outputs a set of tuples $(t_s, t_e, \psi, c)$ where $t_s$ and $t_e$ are the start and end frames of an action, $c$ is the action label, and $\psi$ is the activity score.

\subsection{Snippet-Level Classification}
In our proposed HAM-Net, as illustrated in Figure~\ref{fig:full_model}, for each video, we first divide it into non-overlapping snippets to extract snippet-level features. Using a snippet-level representation rather than a frame-level representation allows us to use existing 3D convolutional features extractors that can effectively model temporal dependencies in the video. Following the two-stream strategy \cite{i3d, feichtenhofer2016convolutional} for action recognition, we extract snippet-level features for both the RGB and flow streams, denoted as ${{\bf x}_{i}^{\text{RGB}} \in \mathbb{R}^D}$ and ${{\bf x}_{i}^{\text{Flow}} \in \mathbb{R}^D}$ respectively. We concatenate both streams to obtain full snippet features ${{\bf x}_{i} \in \mathbb{R}^{2D}}$ for the $i$-th snippet, resulting in a high-level representation of the snippet feature that contains both appearance and motion cues.  

To determine the temporal locations of all activities in the video, we calculate the snippet-level classification scores from classification branch, which is a  convolutional neural network that outputs the class logits commonly known as Class Activation Sequences (CAS) \cite{shou2018autoloc}. We denote the snippet level CAS for all classes for the $i$-th snippet as $\mathbf{s}_i \in \mathbb{R}^{c+1}$. Here, the $c+1$-th class is the background class. 
Since we only have the video-level class scores as ground truth, we need to pool the snippet-level scores $\mathbf{s}_i$ to obtain video-level class scores. There are several pooling strategies in the literature to obtain video-level scores from snippet level scores. We adopt the top-k strategy \cite{islam2020weakly, wtalc} in our setting.  Specifically,  the temporal pooling is performed by aggregating the top-k values from the temporal dimension for each class:
\begin{equation}\label{eq:tpool}
    v_j =  \max_{\substack{
    l \subset \{1,2,\ldots,T \} \\
    |l| = k
    }} \frac{1}{k} \sum_{i \in l}  \mathbf{s}_{i}(j)
\end{equation}

Next, we calculate the video-level class scores by applying softmax operations along the class dimension:
\begin{equation}\label{eq:vid_score}
    p_j = \frac{\exp(v_j)}{\sum_{j'=1}^{c+1} \exp(v_j')}
\end{equation}
where $j=1,2,\ldots,c+1$.

The base classification loss is calculated as the cross entropy loss between the ground-truth video-level class scores $\mathbf{y}$ and the predicted scores $\mathbf{p}$: 
\begin{equation}\label{eq:L_base_fb}
    \mathcal{L}_{\text{BCL}} = -\sum_{j=1}^{c+1} y_j \log(p_j)
\end{equation}

Note that every untrimmed video contains some background portions where there are no actions occurring. These background portions are modeled as a separate class in the classification branch. Hence, the ground-truth background class $y_{c+1}=1$ in Eqn.~\ref{eq:L_base_fb}. One major issue of this approach is that there are no negative samples for the background class, and the model cannot learn background activity by only optimizing with positive samples. To overcome this issue, we propose a hybrid attention mechanism in the attention branch to further explore the ``action-ness'' score of each segment. 




\subsection{A Hybrid Attention Mechanism for Weak Supervision}
To suppress background classes from the video, we incorporate an attention module to differentiate foreground and background actions following the background modeling strategy in several weakly-supervised action detection papers \cite{nguyen2019weaklybackgroundmodeling, lee2020backgroundbasnet,  liu2019weaklycleannet}. The goal is to predict an attention score for each snippet that is lower in the frames where there is no activity instance (i.e.,~background activity) and higher for other regions. Although the classification branch predicts the probability of background action in the snippets, a separate attention module is more effective to differentiate between the foreground and background classes for several reasons. First, most of the actions in a video occur in regions where there are high motion cues; the attention branch can initially detect the background region only from motion features. Second, it is easier for a network to learn two classes (foreground vs.~background) rather than a large number of classes with weak supervision. 

\subsubsection{Soft Attention Score} 

The input to the attention module is the snippet-level feature $x_i$, and it returns a single foreground attention score $a_i$:

\begin{equation}
    a_i = g({\bf x}_i; \Theta),
\end{equation}
where $a_i \in [0, 1]$, and $g(\cdot; \Theta)$ is a function with parameters $\Theta$ that is designed with two temporal convolution layers followed by a sigmoid activation layer.

To create negative samples for the background class, we multiply the snippet level class logit (i.e.,~CAS) $s_i(j)$ for each class $j$ with the snippet-level attention score $a_i$ for the $i$-th snippet, and obtain attention-guided snippet-level class scores $s^{\text{attn}}_{i}(j) = s_i(j) \otimes a_i$, where $\otimes$ is the element-wise product. $s^{\text{attn}}$ serves as a set of snippets without any background activity, which can be considered as negative samples for the background class. Following Eqns.~\ref{eq:tpool} and ~\ref{eq:vid_score}, we obtain video level attention-guided class scores $p^{\text{attn}}_{j}$ for class label $j$: 


\begin{equation}
    v^{\text{attn}}_{j} =  \max_{\substack{
    l \subset \{1,2,\ldots,T \} \\
    |l| = k
    }} \frac{1}{k} \sum_{i \in l}  \mathbf{s}^{\text{attn}}_{ i}(j)
\end{equation}

\begin{equation}
    p^{\text{attn}}_j = \frac{\exp(v^{\text{attn}}_{j})}{\sum_{j'=1}^{c+1} \exp(v^{\text{attn}}_{j'})}
\end{equation}
where $j=1,2,\ldots,c+1$. Note that $p^{\text{attn}}_j$ does not contain any background class, since  the background class has been suppressed by the attention score $a_i$. From $p^{\text{attn}}_j$, we calculate the soft attention-guided loss (SAL) function

\begin{equation}\label{eq:L_base_f}
    \mathcal{L}_{\text{SAL}} = -\sum_{j=1}^{c+1} y_{j}^f \log(p_j )
\end{equation}

Here, $y^f_{j}$ contains only the foreground activities, i.e., the background class $y_{c+1}^f=0$, since the attention score suppresses the background activity. 


\subsubsection{Semi-Soft Attention Score}
Given the snippet-level class score $s_i$ and soft-attention score $a_i$ for the  $i$-th snippet, we calculate the semi-soft attention scores by thresholding the soft attention $a_i$ by a  particular value $\gamma \in [0,1]$,

\begin{align*}
    {a}^{\text{semi-soft}}_{i} &= \begin{cases}
      a_i, & \text{if} \ a_i < \gamma \\
      0, & \text{otherwise}
    \end{cases} 
\end{align*}

Note that the semi-soft attention ${a}^{\text{semi-soft}}_{i}$ both drops the most discriminative regions and attends to the foreground snippets only; hence, the semi-soft attention guided video-level class scores will only contain foreground activities. This design helps to better model the background, as discussed in the ablation studies section. Denote the video-level class scores associated with semi-soft attention as $p^{\text{semi-soft}}_{ j}$, where $j=1,2,\ldots,c+1$. We calculate the semi-soft attention loss:

\begin{equation} \label{eq:L_drop_fb}
     \mathcal{L}_{\text{SSAL}} = -\sum_{j=1}^{c+1} y_j^f \log(p^{\text{semi-soft}}_{ j})
\end{equation}
where $y^f$ is the ground truth label without background activity, i.e., $y^f_{c+1}=0$, since the semi-soft attention suppresses the background snippets along with removing the most discriminative regions.

\subsubsection{Hard Attention Score} In contrast to semi-soft attention, hard attention score is calculated by
\begin{align*}
    {a}^{\text{hard}}_{i} = \begin{cases}
      1, & \text{if} \ a_i < \gamma \\
      0, & \text{otherwise}
    \end{cases}
\end{align*}
With hard attention score, we obtain another set of video-level class scores by multiplying them with the original snippet-level logit $s_i(j)$ and temporally pooling the scores following Eqn.~\ref{eq:tpool} and Eqn.~\ref{eq:vid_score}. We obtain the  hard-attention loss:
\begin{equation} \label{eq:L_drop_f}
     \mathcal{L}_{\text{HAL}} = -\sum_{j=1}^{c+1} y_j \log(p^{\text{hard}}_{ j})
\end{equation}
where $y$ is the ground truth label with background activity, i.e., $y_{c+1}=1$, since the hard attention does not suppress the background snippets, rather, it only removes the more discriminative regions of a video. 

\subsubsection{Loss Functions}
Finally, we train our proposed HAM-Net using the following joint loss function:
\begin{equation}
  \begin{split}
      \mathcal{L} &= \lambda_0  \mathcal{L}_{\text{BCL}} + \lambda_1 \mathcal{L}_{\text{SAL}} + \lambda_2  \mathcal{L}_{\text{SSAL}} + \lambda_3 \mathcal{L}_{\text{HAL}} \\
     &+ \alpha  \mathcal{L}_{\text{sparse}} + \beta  \mathcal{L}_{\text{guide}}
  \end{split}
\end{equation}
where $\mathcal{L}_{\text{sparse}}$ is sparse loss, $\mathcal{L}_{\text{guide}}$ is guide loss, and $\lambda_0$, $\lambda_1$, $\lambda_2$, $\lambda_3$, $\alpha$, and $\beta$ are hyper-parameters.


The sparsity loss $\mathcal{L}_{\text{sparse}}$ is based on the assumption that an action is recognizable from a sparse subset of the video segments \cite{stpn}. The sparsity loss is calculated as the L1-norm of the soft-attention scores:

\begin{equation}
    \mathcal{L}_{\text{sparse}} = \sum_{i=1}^{T} |a_i| 
\end{equation}


Regarding the guide loss $\mathcal{L}_{\text{guide}}$, we consider the soft-attention score $a_i$ as a form of binary classification score for each snippet, where there are only two classes, foreground and background, the probabilities of which are captured by $a_i$ and $1-a_i$ . Hence, $1-a_i$ can be considered as the probability of the $i$-th snippet containing background activity. On the other hand, the background class is also captured by the class activation logits $\mathbf{s}_i (\cdot) \in \mathbb{R}^{c+1}$. To guide the background class activation to follow the background attention, we first  calculate the probability of a particular segment being background activity,
\begin{equation}
    \bar{s}_{c+1} = \frac{\exp(s_{c+1})} {\sum_{j=1}^c \exp(s_j)}
\end{equation}
and then add a guide loss so that the absolute difference between the background class probability and the background attention is minimized: 
\begin{equation}
    \mathcal{L}_{\text{guide}} = \sum_{i=1}^T |1 - a_i - \bar{s}_{c+1}|
\end{equation}

\subsection{Temporal Action Localization}

For temporal localization, we first discard classes which have video-level class score less than a particular threshold (set to $0.1$ in our experiments). For the remaining classes, we first discard the background snippets by thresholding the soft attention scores $a_i$ for all snippets $i$, and obtain class-agnostic action proposals by selecting the one-dimensional connected components of the remaining snippets. Denote the candidate action locations as $\{(t_s, t_e, \psi, c)\}$, where $t_s$ is the start time, $t_e$ is the end time, and $\psi$ is the classification score for class $c$. We calculate the classification score following the outer-inner score of AutoLoc \cite{shou2018autoloc}. Note that for calculating class-specific scores, we use the attention-guided class logits $s^{\text{attn}}_{c}$,


\begin{align}
    & \psi = \psi_{\text{inner}} - \psi_{\text{outer}} + \zeta  p_c^{\text{attn}} \\
    & \psi_{\text{inner}} = \text{Avg}({s}^{\text{attn}}_{c}(t_s: t_e)) \\
    & \psi_{\text{outer}} = \text{Avg}({s}^{\text{attn}}_{c}(t_s-l_m: t_s) +  {s}^{\text{attn}}_{c}(t_e: t_e+l_m))
\end{align}
where $\zeta$ is a hyper-parameter, $l_m = (t_e-t_s)/4$, $p_c^{\text{attn}}$ is the video-level score for class $c$, and ${s}^{\text{attn}}_{c}(\cdot)$ is the snippet-level class logit for class $c$. We apply different thresholds for obtaining action proposals, and remove the overlapping segments with non-maximum suppression.

\section{Experiments} \label{sec:experiments}

\subsection{Experimental Settings}

\subsubsection{Datasets}
We evaluate our approach on two popular action localization datasets: THUMOS14 \cite{THUMOS14} and ActivityNet1.2 \cite{activitynet}. \textbf{THUMOS14} contains 200 validation videos for training and 213 testing videos for testing with 20 action categories. This is a challenging dataset with around 15.5 activity segments and 71\% background activity per video. \textbf{ActivityNet1.2} dataset contains 4,819 videos for training and 2,382 videos for testing with 200 action classes. It contains around 1.5 activity instances (10 times sparser than THUMOS14) and 36\% background activity per video. 


\subsubsection{Evaluation Metrics}
For evaluation, we use the standard protocol and report mean Average Precision (mAP) at various intersection over union (IoU) thresholds. The evaluation code provided by ActivityNet \cite{activitynet} is used to calculate the evaluation metrics.

\subsubsection{Implementation Details}
For feature extraction, we sample the video streams into non-overlapping 16 frame chunks for both the RGB and the flow stream. Flow streams are created using the TV-L1 algorithm \cite{TVL1}. We use the I3D network \cite{i3d} pretrained on the Kinetics dataset \cite{kinetics} to extract both RGB and flow features, and concatenate them to obtain 2048-dimensional snippet-level features. During training we randomly sample 500 snippets for THUMOS14 and 80 snippets for ActivityNet, and during evaluation we take all the snippets.
The classification branch is designed as two temporal convolution layers with kernel size 3, each followed by LeakyReLU activation, and a final linear fully-connected layer for predicting class logits. The attention branch consists of two temporal convolution layers with kernel size 3 followed by a sigmoid layer to predict attention scores between 0 and 1.

We use the Adam \cite{Kingma2015AdamAM} optimizer with learning rate 0.00001, and train for 100 epochs for THUMOS14 and 20 epochs for ActivityNet. For THUMOS14, we set $\lambda_0=\lambda_1=0.8$, $\lambda_2=\lambda_3=0.2$, $\alpha=\beta=0.8$, $\gamma=0.2$, and $k=50$ for top-k temporal pooling. For ActivityNet, we set $\alpha=0.5$, $\beta=0.1$, $\lambda_0=\lambda_1=\lambda_2=\lambda_3=0.5$, and $k=4$, and apply additional average pooling to post-process the final CAS. All the hyperparameters are determined from grid search. For action localization, we set the thresholds from 0.1 to 0.9 with a step of 0.05, and perform non-maximum suppression to remove overlapping segments. 

\subsection{Ablation Studies}\label{subsec:ablation}
\begin{figure}[htbp]
    \centering
    \includegraphics[width=0.47\textwidth]{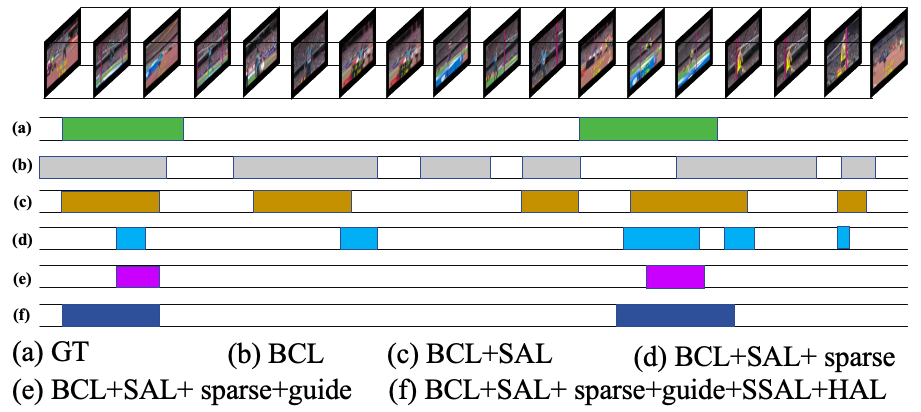}
    \caption{Visualization of the effects of different loss functions on the final localization for a video containing the Long Jump activity. (a) is the ground-truth action location. (b) represents only MIL loss, which predicts many false positives. After adding sparsity and guide loss, in (d) we get rid of those false positives, but still cannot capture full temporal boundaries. (e) shows results from our approach which captures full action boundaries.}
    \label{fig:vis_abl_loss}
    \vspace{-0.2cm}
\end{figure}

We conduct a set of ablation studies on the THUMOS14 dataset to analyze the performance contribution of each component of our proposed HAM-Net. Table \ref{tab:ablation} shows the performance of our method with respect to different loss terms. We use ``AVG mAP" for the performance metric, which is the average of mAP values for different IoU thresholds (0.1:0.1:0.7). The first five experiments are trained without SSAL or HAL loss, i.e., without any temporal dropping mechanism, which we denote as ``{MIL-only mode}'', and the remaining experiments trained with those losses are denoted as ``{MIL and Drop mode}''. Figure~\ref{fig:vis_abl_loss} shows the localization prediction of different experiments on a representative video. Our analysis shows that all the loss components are required to achieve the maximum performance.

\begin{table}[ht]
\fontsize{9}{9}\selectfont
    \centering
    \begin{tabular}{c  c c c c c c c c  c}
    \hline
      \multirow{2}{*}{$\mathcal{L}_{\text{BCL}}$}&
      \multirow{2}{*}{$\mathcal{L}_{\text{SAL}}$}&
      \multirow{2}{*}{$\mathcal{L}_{\text{HAL}}$}&
      \multirow{2}{*}{$\mathcal{L}_{\text{SSAL}}$}&
  \multirow{2}{*}{$\mathcal{L}_{\text{sparse}}$} &       \multirow{2}{*}{$\mathcal{L}_{\text{guide}}$} &

         {AVG}\\
         &&&&&& mAP  \\
    \hline
    \hline
    1) \cmark & - & - & - & - & -  &  24.6 \\
    2) \cmark & \cmark& - & - & - & -  &  30.8 \\
    3) \cmark & \cmark& - & - & - & \cmark  &  28.9  \\
    4) \cmark & \cmark& - & - & \cmark & -  &  30.9  \\
    5) \cmark & \cmark& - & - & \cmark & \cmark &  \bf34.8 \\
    \hline
    6) \cmark & \cmark & \cmark & \cmark &- & -  & 30.9 \\
    7) \cmark & \cmark & \cmark & \cmark & - & \cmark &  31.1  \\
    8) \cmark & \cmark & \cmark & \cmark & \cmark & - & 37.9  \\
    9) \cmark & \cmark & \cmark & - & \cmark & \cmark & 36.6 \\
    10) \cmark & \cmark & - & \cmark & \cmark & \cmark & 38.1 \\
    11) \cmark & \cmark & \cmark & \cmark & \cmark & \cmark &  \bf39.8 \\
    \hline
    \end{tabular}
    \caption{Ablation study on the effectiveness of different combination of loss functions in the localization performance on THUMOS14 in terms of mAP. Here, AVG mAP means the average of mAP values from IoU thresholds 0.1 to 0.7. Adding ${\mathcal{L}_{\text{AL}}}$ in the total loss function improves the mAP from 34.8 to 39.8.}
    \label{tab:ablation}
\end{table}

    \subsubsection{Importance of sparsity and guide loss} Table~\ref{tab:ablation} shows that both sparsity and guide loss are important to acheive better performance. Specifically, in ``MIL-only mode'', adding both sparsity and guide loss provides 4\% mAP gain, and in ``MIL and Drop mode'', the mAP gain is 9\%, suggesting that these losses are more important in ``MIL and Drop mode''. Note that for SSAL and HAL, the discriminativeness of a snippet is measured by the soft-attention scores that are learned by sparsity and guide loss.  Without sparsity loss, the majority of the soft-attention scores remain close to 1, making the snippet dropping strategy ineffective. Moreover, the guide loss itself does not increase the localization performance significantly  without the sparsity loss (experiment 3 and experiment 7 in Table~\ref{tab:ablation}); however, combined with sparsity loss it shows the best performance improvement (experiment 5 and experiment 11 in Table~\ref{tab:ablation}).
    
    \subsubsection{Importance of attention losses} We observe that the attention losses can significantly improve the performance. Table~\ref{tab:ablation} shows that only incorporating $\mathcal{L}_{\text{SAL}}$ achieves 6.2\% average mAP gain over the BCL-only model. From experiment-9 and experiment-10 in Table~\ref{tab:ablation}, we see that both HAL and SSAL individually improve the performance, and we get the best performance when we combine them. Specifically, the combination of HAL and SSAL improves the performance by 5\% over the best score in ``MIL-only mode''. Figure \ref{fig:qual_examples} shows visualization examples of the effectiveness of the losses on a representative video. We can observe that the MIL-only model fails to capture several parts of a full action instance (i.e.~, Long Jump). Incorporating attention losses helps to capture the action in its entirety. 
    
    \subsubsection{Importance of dropping snippets by selective thresholding} 
\begin{figure}[h!]
\begin{center}
\begin{subfigure}{0.23\textwidth}
\includegraphics[width=0.98\linewidth]{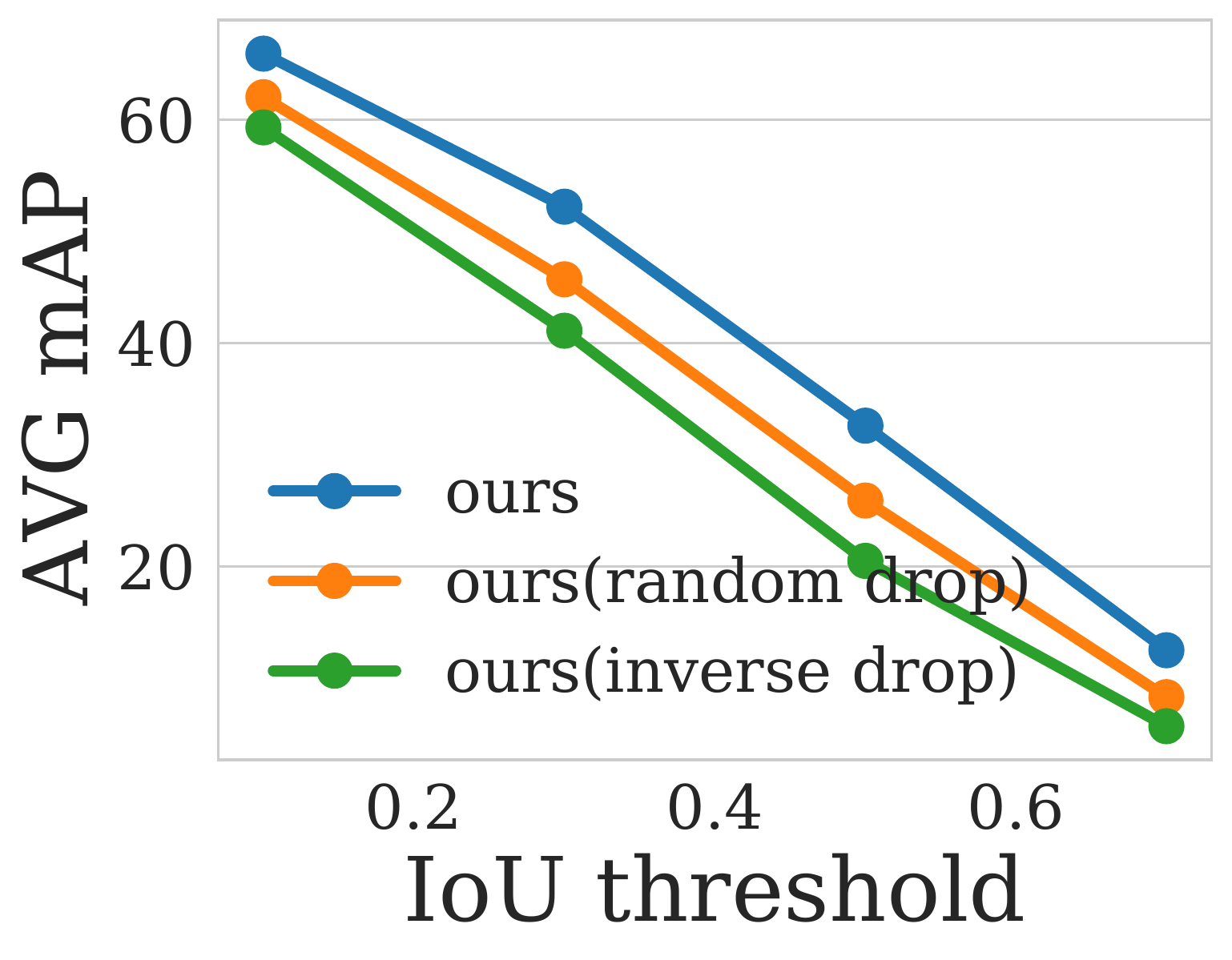}
\vspace{-0.2cm}
\caption{}
\label{fig:sub_base}
\end{subfigure}
\begin{subfigure}{0.23\textwidth}
\includegraphics[width=0.98\linewidth]{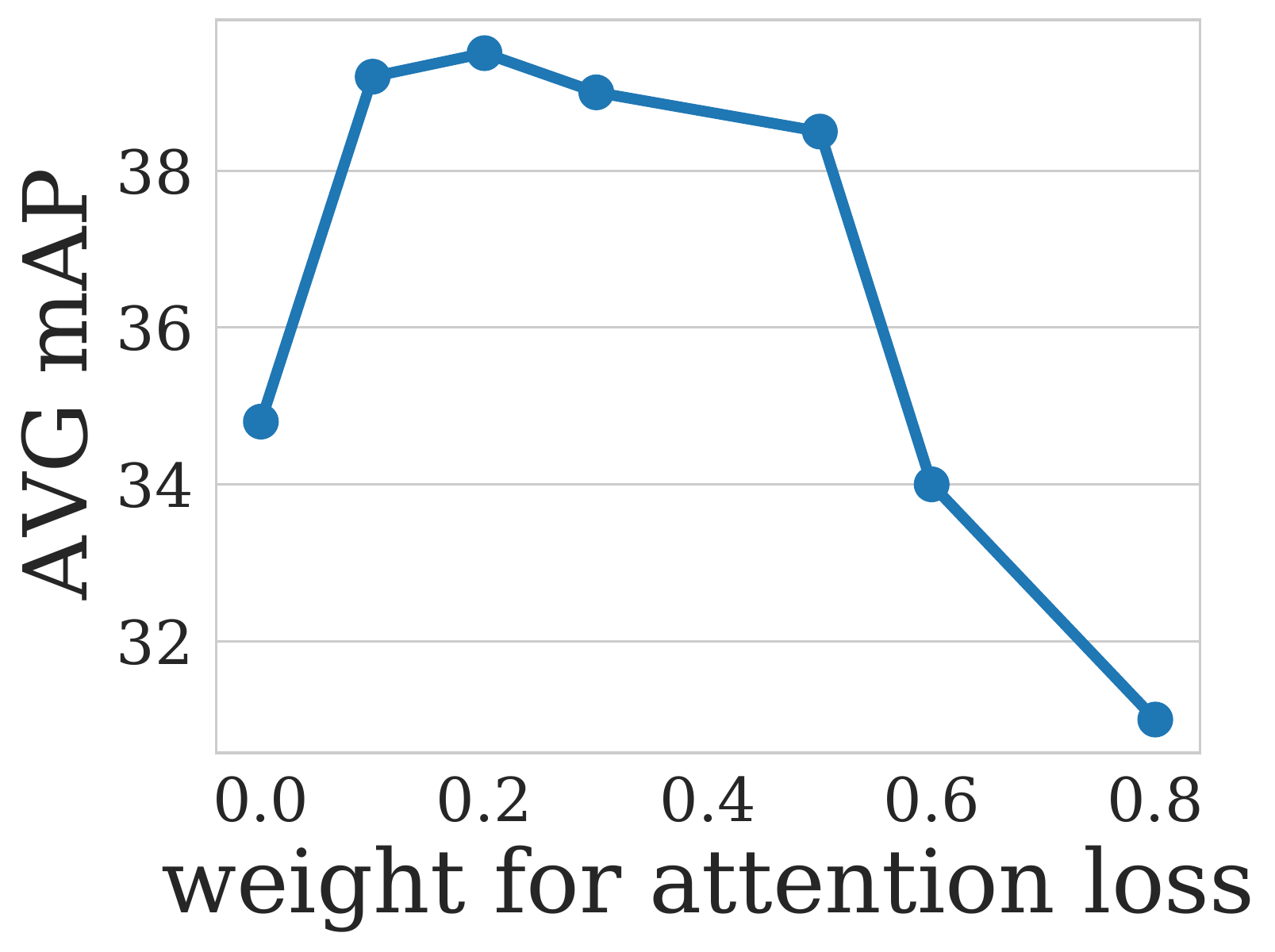}
\vspace{-0.2cm}
\caption{}
\label{fig:sub_l2}
\end{subfigure}
\end{center}
    \vspace{-0.40cm}
    \caption{(a) Ablation study on the importance of dropping snippets by selective thresholding. Other approaches like random dropping or inverse selective thresholding do not work well. (b) Ablation study on the importance of SSAL and HAL. A lower weight causes the model to learn only the most distinctive parts, and a higher weight gives too much focus to the less distinctive parts.}
    \label{fig:ablation_hp}
    \vspace{-0.3cm}
\end{figure}

For calculating the HAL loss and the SSAL loss, we drop the more discriminative parts of the video and train on the less discriminative parts, assuming that focusing on less discriminative parts will help the model to learn the  completeness of actions. To confirm our assumption here, we create two baselines: ``\textbf{ours with random drop}'' where we randomly drop video snippets, similar to Hide-and-Seek \cite{Singh2017HideandSeekFA}, and ``\textbf{ours with inverse drop}'', where we drop the less discriminative parts instead of dropping the most discriminative parts. We show the performance comparison between these models in Figure~\ref{fig:sub_base}. Results show that randomly dropping snippets is slightly more effective than the baseline, and dropping the less discriminative parts decreases the localization performance. Our approach performs much better than randomly dropping snippets or dropping less discriminative snippets, which proves the efficacy of selectively dropping more discriminative foreground snippets. 

    \subsubsection{Ablation on $\lambda_2$ and $\lambda_3$} For this analysis, we set $\lambda_2=\lambda_3=\lambda$. In Figure~\ref{fig:sub_l2}, we analyze the effect of  $\lambda$ to the performance. Note that $\lambda=0$ denotes ``MIL only mode", which achieves an average mAP of 34.8\%. Increasing the value of $\lambda$ results in performance improvement until $\lambda$ reaches 0.2, after which we observe performance degradation. The reason is that a lower weight does not incorporate $\mathcal{L}_{\text{SSAL}}$ and $\mathcal{L}_{\text{HAL}}$ effectively during training. On the contrary, a higher weight provides too much importance on the less discriminative parts, which might cause the model to ignore the more discriminative regions in every iteration, resulting in poor localization performance. The optimum value of 0.2 balances out both of the issues. 
    

\begin{table*}[htbp!]
\fontsize{9}{9}\selectfont
    \centering
    \begin{tabular}{c c | c||r r r r r r r | r}
    \hline
    \multirow{2}{*}{} & \multirow{2}{*}{\bf Method} & \multirow{2}{*}{\bf Feature} & \multicolumn{8}{c}{\bf IoU}\\ 
    \cline{4-11}
    & & & \bf0.1 & \bf0.2 &  \bf0.3 & \bf0.4 & \bf0.5 & \bf0.6 & \bf0.7 & \bf AVG \\
    \hline
    \hline

    \multirow{7}{*}{} 
    & R-C3D \cite{Xu2017RC3DRC} & -  & 54.5 & 51.5 & 44.8 & 35.6 & 28.9 &- &- & -\\
    & SSN \cite{ssn} & -  & 66.0 & 59.4 & 51.9 & 41.0 & 29.8 & - & - & - \\
    & BSN \cite{lin2018bsn} & - & - & - & 53.5 & 45.0 & 36.9 & 28.4 & 20.0 & - \\
    & G-TAD \cite{xu2019gtad} & - & - & - & 54.5 & 47.6 & 40.2 & \bf30.8 & \bf23.4 & -\\
    & P-GCN \cite{Zeng_2019_ICCV} & -  & \bf69.5 & \bf67.8 & \bf63.6 & \bf57.8 & \bf49.1 &  - & - & -\\
    \hline
    \multirow{16}{*}{} 
    & Hide-and-Seek \cite{Singh2017HideandSeekFA} & -  & 36.4 & 27.8 & 19.5 & 12.7 & 6.8 & - & -& -\\
    & UntrimmedNets \cite{wang2017untrimmednets} & -  & 44.4 & 37.7 &28.2 & 21.1 & 13.7 & -& -& -\\
    & STPN \cite{stpn} & I3D  & 52.0 & 44.7 & 35.5 & 25.8 & 16.9 & 9.9 & 4.3 & 26.4 \\
    & AutoLoc \cite{shou2018autoloc} & UNT  & - & - & 35.8 & 29.0 & 21.2 & 13.4 & 5.8 & -\\
    & W-TALC \cite{wtalc} & I3D & 55.2 & 49.6 & 40.1 & 31.1 & 22.8 &-& 7.6& -\\
    & Liu \etal~\cite{liu2019completeness} & I3D  & 57.4 & 50.8 & 41.2 & 32.1 & 23.1 & 15.0 & 7.0 & 32.4 \\
    & MAAN \cite{yuan2018maan} & I3D & 59.8 & 50.8 & 41.1 & 30.6 & 20.3 & 12.0 & 6.9 & 31.6\\
    & TSM \cite{yu2019weaktsm}& I3D & - & - & 39.5 & - & 24.5 & - & 7.1& -\\
    & CleanNet \cite{liu2019weaklycleannet} & UNT & - & - & 37.0 & 30.9 & 23.9 & 13.9 & 7.1 & -\\
    & 3C-Net \cite{narayan20193cnet} & I3D & 56.8 & 49.8 & 40.9 & 32.3 & 24.6 & - & 7.7 & -\\
    & Nguyen \etal~\cite{nguyen2019weaklybackgroundmodeling} & I3D & 60.4 & 56.0 & 46.6 & 37.5 & 26.8 & 17.6 & 9.0 & 36.3 \\
    & WSGN \cite{fernando2020weaklywsgn} & I3D & 57.9 & 51.2 & 42.0 & 33.1 & 25.1 & 16.7 & 8.9 & 33.6\\
    & Islam \etal~\cite{islam2020weakly} & I3D & 62.3 & - & 46.8 & - & 29.6 & - & 9.7 & -\\
    & BaS-Net \cite{lee2020backgroundbasnet} & I3D & 58.2 & 52.3 & 44.6 & 36.0 & 27.0 & 18.6 & 10.4 & 35.3 \\
    & DGAM \cite{Shi_2020_CVPR_attention_modeling} & I3D & 60.0 & 54.2 & 46.8 & 38.2 & 28.8 & 19.8 & 11.4 & 37.0 \\ 
    & \bf{HAM-Net (Ours)} & I3D & \bf65.4 & \bf59.0 & \bf50.3 &  \bf41.1 & \bf31.0 &  \bf20.7 & \bf11.14 & \bf39.8 \\ 
    \hline
    \end{tabular}
    \caption{Comparison of our algorithm with other state-of-the-art methods on the THUMOS14 dataset for temporal action localization.}
    \label{tab:thumos_result}
    \vspace{-0.3cm}
\end{table*}

\subsection{Performance Comparison to State-of-the-Art}
Table~\ref{tab:thumos_result} summarizes performance comparisons between our proposed HAM-Net and state-of-the-art fully-supervised and weakly-supervised TAL methods on the THUMOS14 dataset. We report mAP scores at different IoU thresholds. `AVG' is the average mAP for IoU 0.1 to 0.7 with step size of 0.1. With weak supervision, our proposed HAM-Net achieves state-of-the-art scores on all IoU thresholds. Specifically, HAM-Net achieves 2.2\% more mAP than the current best scores at IoU threshold 0.5. Moreover, our HAM-Net outperforms some fully-supervised TAL models, and even shows comparable results with some recent fully-supervised TAL methods. 

In Table~\ref{tab:anet_result}, we evaluate HAM-Net on the ActivityNet1.2 dataset. HAM-Net outperforms other WTAL approaches on ActivityNet1.2 across all metrics, verifying the effectiveness of our proposed HAM-Net.

\begin{table}[h!]
    \fontsize{9}{9}\selectfont
    \centering
    \begin{tabular}{c|c||r  r  r  | r}
    \hline
    \multirow{2}{*}{} & \multirow{2}{*}{\bf Method} & \multicolumn{4}{c}{\bf IoU}\\ 
    \cline{3-6}
    && \bf0.5 &  \bf0.75 &  \bf0.95 & \bf{AVG} \\
    \hline
    \hline
     \bf Full & SSN & 41.3 & 27.0 & 6.1 & 26.6 \\ 
    \hline
    \multirow{11}{*}{\bf Weak} 
    & UntrimmedNets  & 7.4  & 3.2  & 0.7 & 3.6 \\ 
    &  AutoLoc& 27.3 & 15.1 &  3.3 & 16.0 \\ 
    & W-TALC & 37.0  & 12.7 & 1.5 & 18.0 \\ 
    & Islam \etal& 35.2 & - &  - & - \\ 
    & TSM & 28.3  & 17.0 & 3.5 & 17.1 \\ 
    & 3C-Net& 35.4 & - & - & 21.1 \\ 
    & CleanNet & 37.1 & 20.3  & 5.0 & 21.6 \\ 
    & Liu \etal & 36.8 &  22.0 &  5.6 & 22.4 \\ 
    & BaS-Net & 34.5 & 22.5 & 4.9 & 22.2 \\ 
    & DGAM& \bf41.0 & 23.5  & \bf5.3 & 24.4\\ 
    & \bf{HAM-Net (Ours)} & \bf41.0 &  \bf24.8 &  \bf5.3 & \bf25.1 \\
    \hline
    \end{tabular}
    \caption{Comparison of our algorithm with other state-of-the-art methods on the ActivityNet1.2 validation set for temporal action localization. AVG means average mAP from IoU 0.5 to 0.95 with 0.05 increment.}
    \label{tab:anet_result}
    \vspace{-0.3cm}
\end{table}

\subsection{Qualitative Performance}

\begin{figure}[h!]
\centering
        \begin{subfigure}[b]{0.4\textwidth}   
            \includegraphics[width=\textwidth]{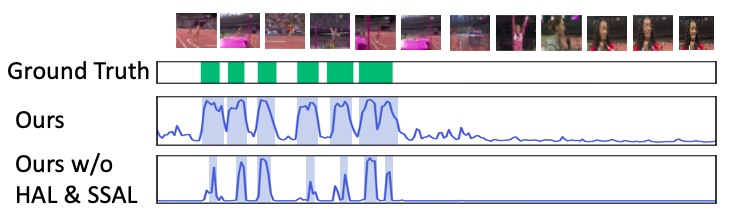}
            \vspace{-0.6cm}
            \caption{High Jump}
            \label{fig:highjump}
    \end{subfigure}

    \begin{subfigure}[b]{0.4\textwidth}
            \includegraphics[width=\textwidth]{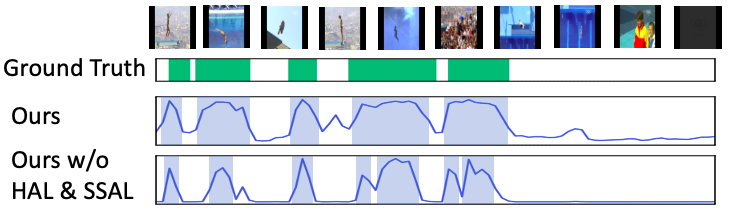}
            \vspace{-0.6cm}
            \caption{Diving}
            \label{fig:diving}
    \end{subfigure}
    \vspace{-0.4cm}
    \caption{Qualitative results on THUMOS14. The horizontal axis denotes time. On the vertical axis, we sequentially plot the ground truth detection, our detection scores, and detection scores of ours without HAL and SSAL. SSAL and HAL helps to learn the full context of an action.}
    \label{fig:qual_examples}
\end{figure}

We show some representative examples in Fig.~\ref{fig:qual_examples}. For each video, the top row shows example frames, the next row represents ground-truth localization, ``Ours'' is our prediction, and ``Ours w/o HAL \& SSAL'' is ours trained without $\mathcal{L}_{\text{HAL}}$ and $\mathcal{L}_{\text{SSAL}}$.
Fig.~\ref{fig:qual_examples} shows that our model clearly captures the full temporal extent of activities, while ``ours w/o HAL \& SSAL'' focuses only on the more discriminative snippets. 

\vspace{-0.2cm}
\section{Conclusion}
We presented a novel framework called HAM-Net to learn temporal action localization from only video-level supervision during training. We introduced a hybrid attention mechanism including soft, semi-soft, and hard attentions to differentiate background frames from foreground ones and to capture the full temporal boundaries of the actions in the video, respectively. We perform extensive analysis to show the effectiveness of our approach. Our approach achieves state-of-the-art performance on both the THUMOS14 and ActivityNet1.2 dataset. 

\section{Acknowledgments}
This material is based upon work supported by the U.S.~Department of Homeland Security under Award Number 2013-ST-061-ED0001. The views and conclusions contained in this document are those of the authors and should not be interpreted as necessarily representing the official policies, either expressed or implied, of the U.S. Department of Homeland Security.

\bibliography{main}

\newpage
\clearpage
\subimport{./}{appendix}

\end{document}

%% file: appendix.tex
The supplemental material contains additional experiments, visualization and ablation studies.

\section{Experiments}
\subsection{Action Classification}
Table.~\ref{tab:action_class} shows action classification performance of our approach in comaprison with other state-of-the-arts in THUMOS14 and ActivityNet1.2 dataset. We use classification mean average precision (mAP) for evaluation. We see that the classification performance of our approach is very competitive with the SOTAs, specially in THUMOS14 we achieve 7.2\% mAP improvement over 3C-Net~\cite{narayan20193cnet}. We also achieve very competitive performance in ActivityNet dataset. Although our approach has not been designed for  video action recognition task, it's high performance in action classification reveals the robustness of our method.

\begin{table*}[tbp]
    \centering
    \begin{tabular}{c|c|c}
    \hline
    Methods & THUMOS14 & ActivityNet1.2 \\
    \hline
    iDT+FV~\cite{wang2013action} & 63.1 & 66.5 \\
    C3D~\cite{c3d} & - & 74.1 \\
    TSN~\cite{Wang2016TemporalSN} & 67.7 & 88.8 \\
    W-TALC~\cite{wtalc} & 85.6 & \bf93.2 \\
    3C-Net~\cite{narayan20193cnet} & 86.9 & 92.4 \\
    Ours & \bf94.1 & 90.3 \\
    \hline
    \end{tabular}
    \caption{Action Classification performance of our method with state-of-the-arts methods on THUMOS14 and ActivityNet1.2 dataset in terms of classification mAP. }
    \label{tab:action_class}
\end{table*}

\subsection{Detailed Performance on ActivityNet1.2}
Table.~\ref{tab:anet_result_details} shows detailed performance of our approach on ActivityNet1.2 dataset in terms of localization mAP for different IoU thresholds.  

\begin{table*}[htbp]
    \fontsize{8}{10}\selectfont
    \centering
    \begin{tabular}{c|c||r r r r r r r r r r  | r}
    \hline
    \multirow{2}{*}{\bf Supervision} & \multirow{2}{*}{\bf Method} & \multicolumn{11}{c}{\bf IoU}\\ 
    && \bf0.5 & \bf0.55 & \bf0.6 & \bf0.65 & \bf0.7 & \bf0.75 &  \bf0.8 & \bf0.85 & \bf0.9 & \bf0.95 & \bf{AVG} \\
    \hline
    \hline
     \bf Full & SSN \cite{ssn} & 41.3 & 38.8 & 35.9 & 32.9 & 30.4 & 27.0 & 22.2 & 18.2 & 13.2 & 6.1 & 26.6 \\
    \hline
    \multirow{11}{*}{\bf Weak} 
    & UntrimmedNets \cite{wang2017untrimmednets} & 7.4 & 6.1 & 5.2 & 4.5 & 3.9 & 3.2 & 2.5 & 1.8 & 1.2 & 0.7 & 3.6 \\
    &  AutoLoc \cite{shou2018autoloc} & 27.3 & 24.9 & 22.5 & 19.9 & 17.5 & 15.1 & 13.0 & 10.0 & 6.8 & 3.3 & 16.0 \\
    & W-TALC \cite{wtalc} & 37.0 & 33.5 & 30.4 & 25.7 & 14.6 & 12.7 & 10.0 & 7.0 & 4.2 & 1.5 & 18.0 \\
    & TSM~\cite{yu2019weaktsm} & 28.3 & 26.0 & 23.6 & 21.2 & 18.9 & 17.0 & 14.0 & 11.1 & 7.5 & 3.5 & 17.1 \\
    & 3C-Net~\cite{narayan20193cnet} & 35.4 & -  & - & - & 22.9 & - & - & - & 8.5 & - & 21.1 \\
    & CleanNet~\cite{liu2019weaklycleannet} & 37.1 & 33.4 & 29.9 & 26.7 & 23.4 & 20.3 & 17.2 & 13.9 & 9.2 & 5.0 & 21.6 \\
    & Liu \etal~\cite{liu2019completeness} & 36.8 & - & - & - & - & 22.0 & - & - & - & 5.6 & 22.4 \\
    & Islam \etal~\cite{islam2020weakly} & 35.2 & - & - & - & 16.3 & - & - & - & - & - & - \\
    & BaS-Net~\cite{lee2020backgroundbasnet} & 34.5 & - & - & - & - & 22.5 & - & - & - & 4.9 & 22.2 \\
    & DGAM~\cite{shi2020weaklysupervisedgam} & \bf41.0 & 37.5 & 33.5 & 30.1 & 26.9 & 23.5 & 19.8 & 15.5 & \bf10.8 & \bf5.3 & 24.4\\
    & \bf{Ours} & \bf41.0 & \bf37.9 & \bf34.6 & \bf31.3 & \bf28.1 & \bf24.8 & \bf21.1 & \bf16.0 & \bf10.8 & \bf5.3 & \bf25.1 \\
    \hline
    \end{tabular}
    \caption{Comparison of our algorithm with other state-of-the-art methods on the ActivityNet1.2 validation set for temporal action localization.}
    \label{tab:anet_result_details}
\end{table*}

\section{More Ablation}

\begin{figure*}[!htbp]
\begin{center}
\begin{subfigure}{0.3\textwidth}
\includegraphics[width=0.9\linewidth]{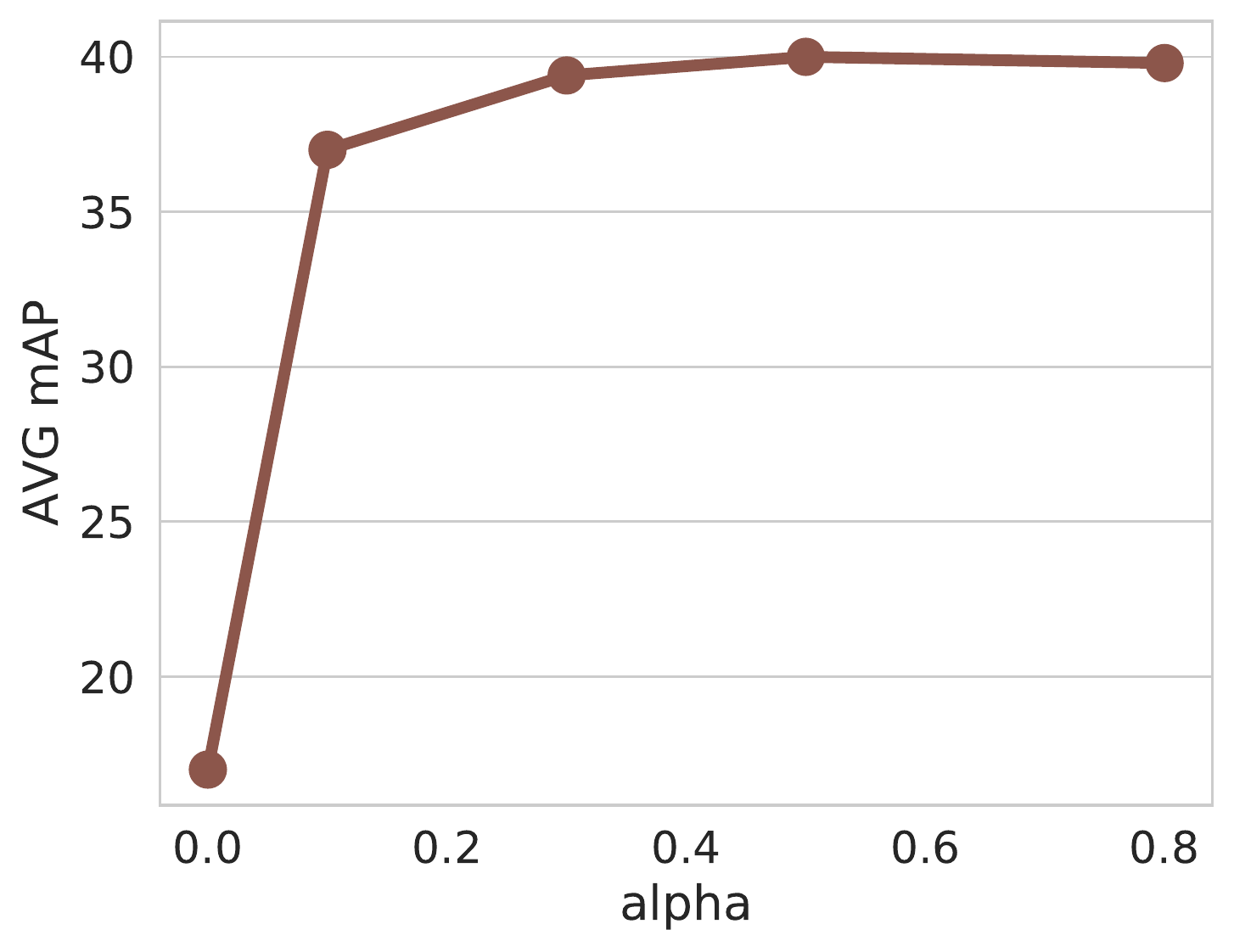}
\caption{}
\label{fig:abl_alpha}
\end{subfigure}
\begin{subfigure}{0.3\textwidth}
\includegraphics[width=0.9\linewidth]{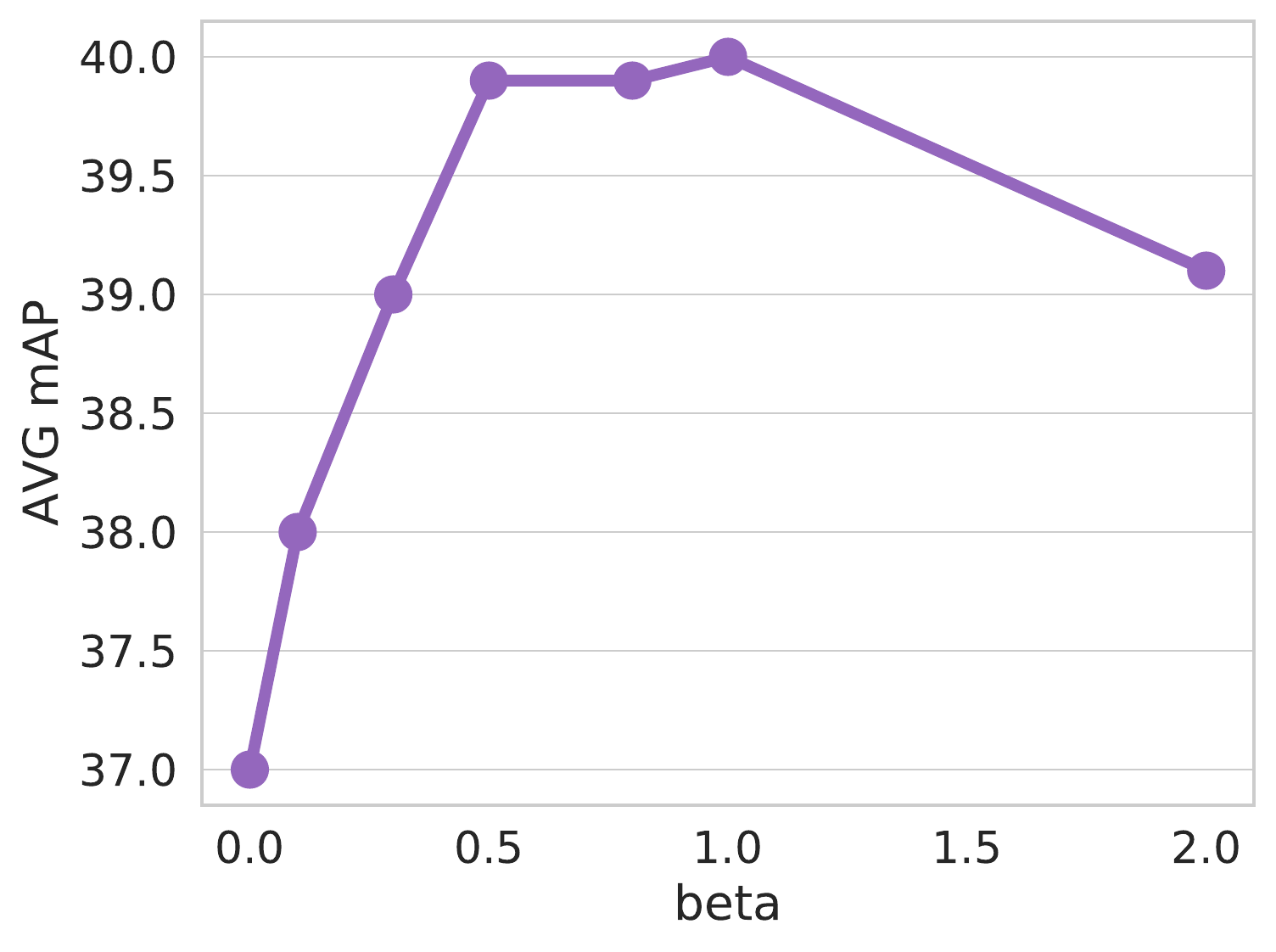}
\caption{}
\label{fig:abl_beta}
\end{subfigure}
\begin{subfigure}{0.3\textwidth}
\includegraphics[width=0.9\linewidth]{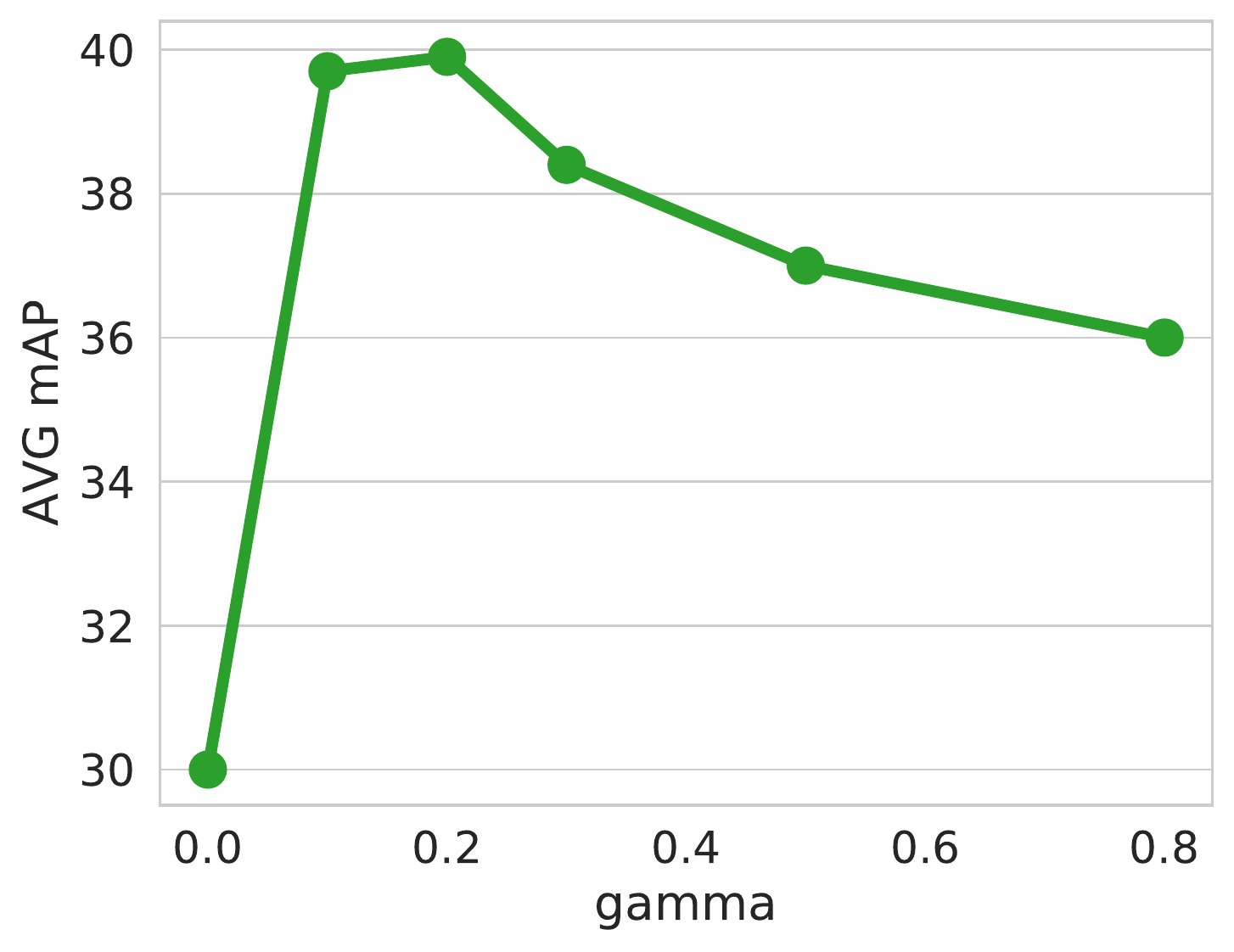}
\caption{}
\label{fig:abl_drop}
\end{subfigure}

\begin{subfigure}{0.3\textwidth}
\includegraphics[width=0.9\linewidth]{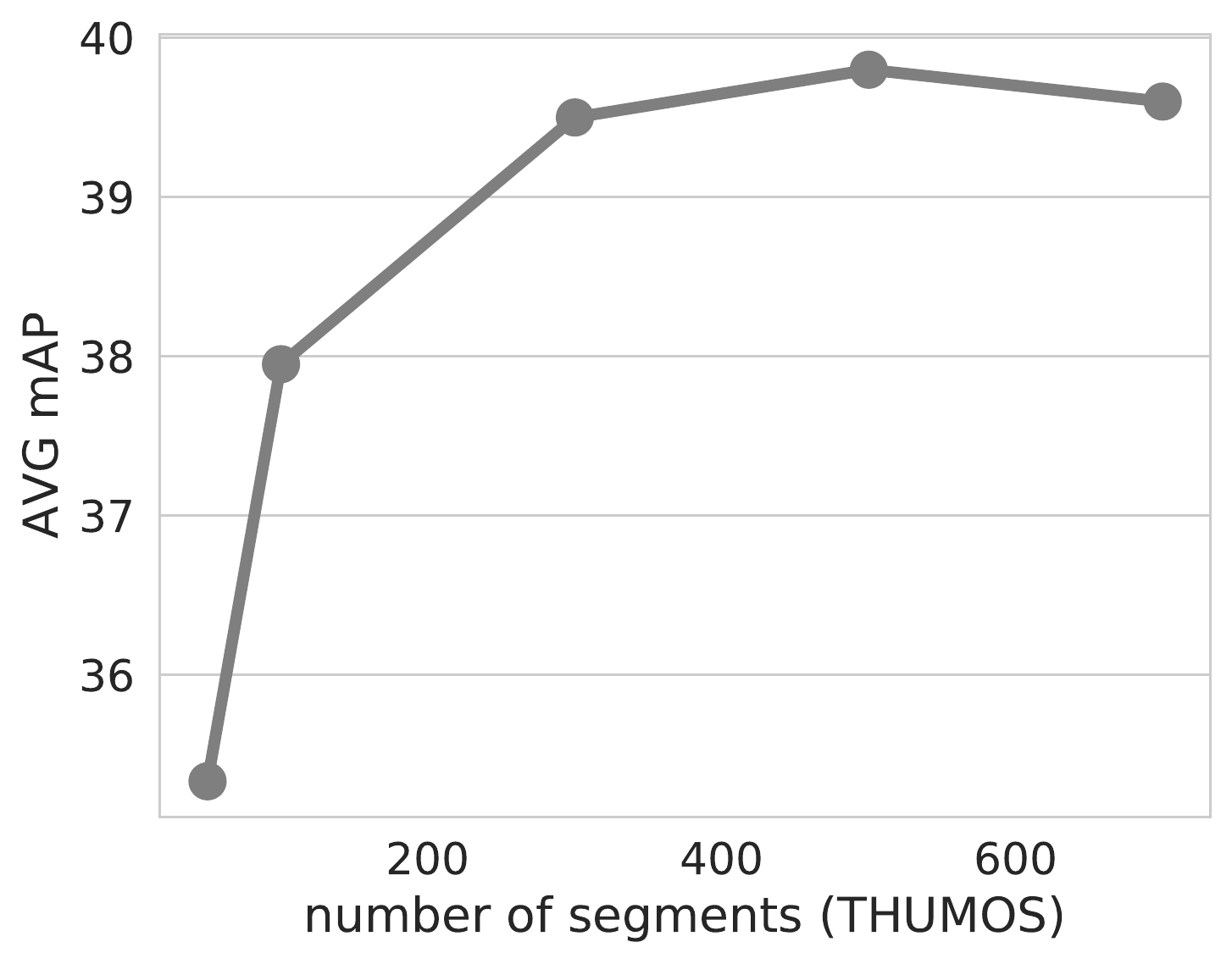}
\caption{}
\label{fig:abl_seg_th}
\end{subfigure}
\begin{subfigure}{0.3\textwidth}
\includegraphics[width=0.9\linewidth]{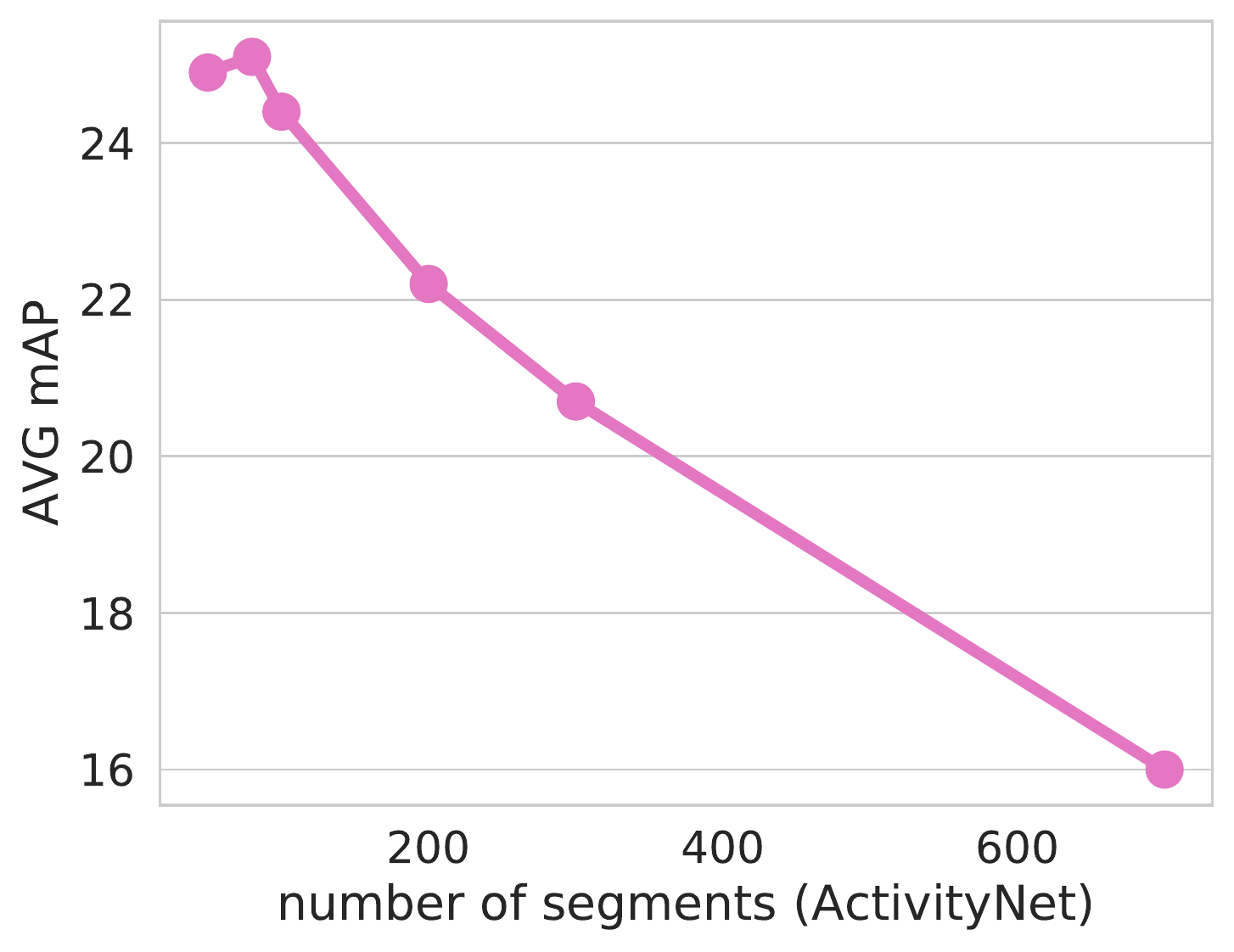}
\caption{}
\label{fig:abl_seg_anet}
\end{subfigure}

\end{center}
    \caption{(a) Ablation on the weight of sparsity loss. (b) Ablation on the weight of guide loss. (c) Ablation on the drop-threshold for dropping snippets in the HAD module. (d) and (e) Ablation on the number of segments for a video during training.}
    \label{fig:ablation_more}
\end{figure*}

Fir.~\ref{fig:ablation_more} shows ablation studies on the hyper-parameters $\alpha$, $\beta$, and drop threshold $\gamma$ on THUMOS14 dataset. AVG mAP is the mean mAP value from IoU threshold 0.1 to 0.7 incremented by 0.1. Fig.~\ref{fig:abl_alpha} shows the performance for different weights on sparsity loss. Without sparsity loss, the model hardly learns any localization. As $\alpha$ increases, localization performance increases as well, and we get the best score for $\alpha=0.8$. Fig.~\ref{fig:abl_beta} reveals the performance improvement for different weights on guide loss. We empirically find that $\beta=0.8$ gives the best performance. In Fig.~\ref{fig:abl_drop}, we see the mAP performance for different values of dropping threshold $\gamma$. 
Fig.~\ref{fig:abl_seg_th} and Fig.~\ref{fig:abl_seg_anet} show the effect of video length during training for THUMOS14 and ActivityNet respectively. Note that THUMOS14 contains more denser videos with a large number of activities per video. Hence we observe that the performance increase for larger video length for THUMOS14, whereas, ActivityNet performs best for 80 length segments. Also, note that the number of segments are chosen randomly only during training. We use all segments during evaluation.

\section{More Qualitative Examples}
\begin{figure*}[htbp]
\label{fig:qual}
\begin{center}
    \begin{subfigure}[b]{0.8\textwidth}            
            \includegraphics[width=\textwidth]{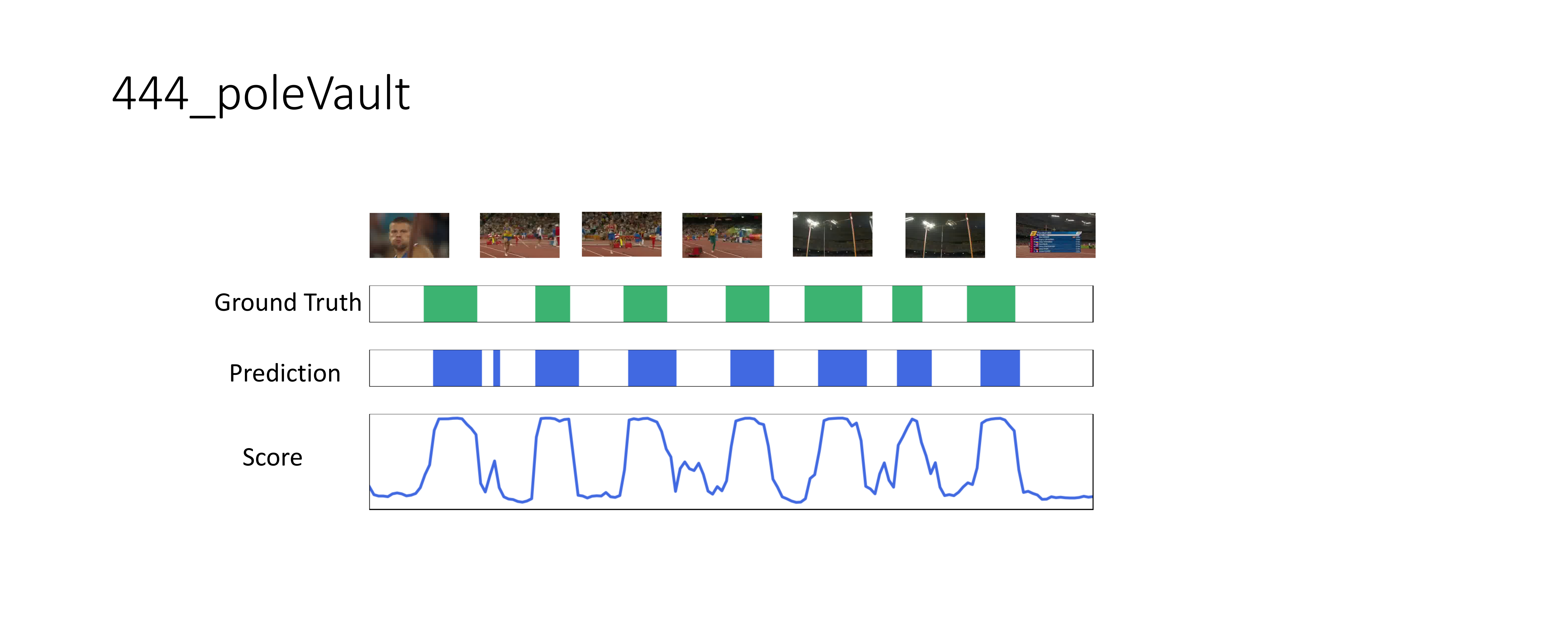}
            \caption{Pole Vault}
            \label{fig:444polevaule}
    \end{subfigure}

    \begin{subfigure}[b]{0.8\textwidth}            
            \includegraphics[width=\textwidth]{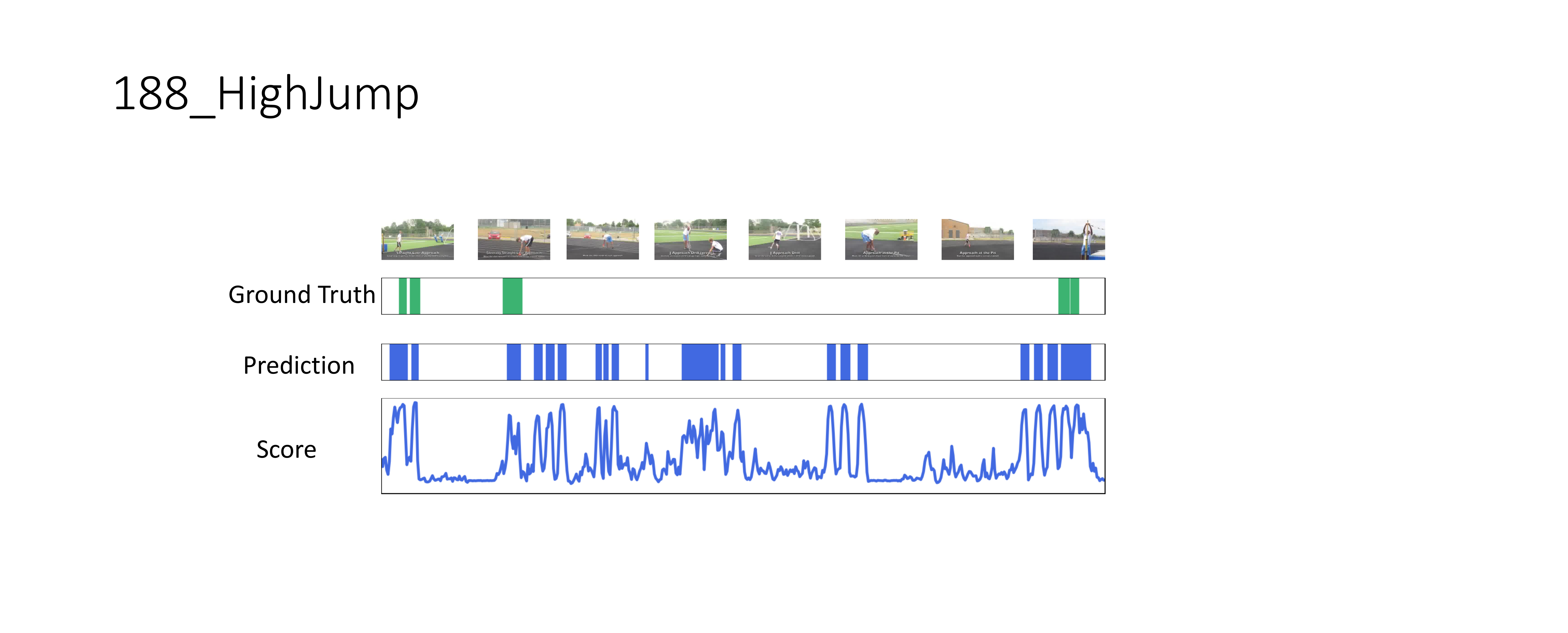}
            \caption{High Jump}
            \label{fig:188highjump}
    \end{subfigure}
    
    \begin{subfigure}[b]{0.8\textwidth}            
            \includegraphics[width=\textwidth]{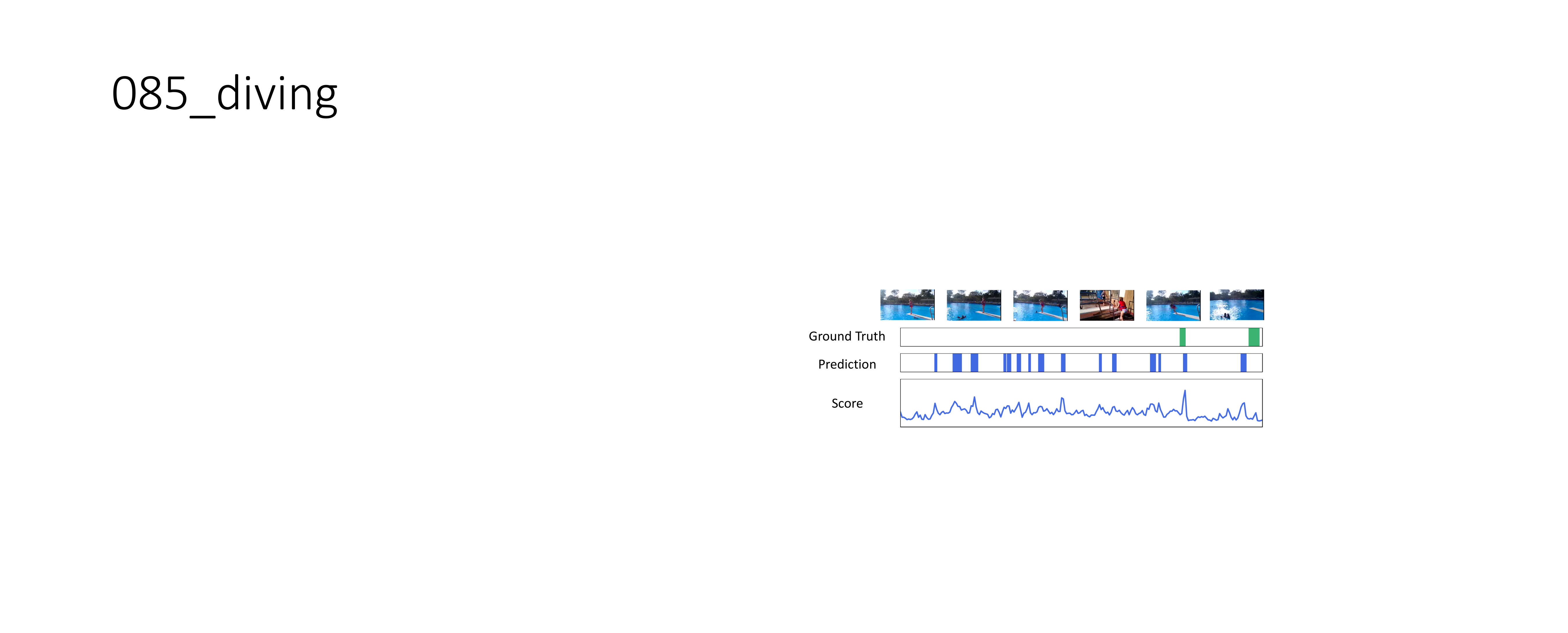}
            \caption{Diving}
            \label{fig:85diving}
    \end{subfigure}

\end{center}
    \caption{Qualitative results on THUMOS14. The horizontal axis denotes time. On the vertical axis, we sequentially plot the ground truth, our predicted localization, and our prediction score. (b) and (c) represent failure examples of our approach. }
    \label{fig:qual_more}
\end{figure*}

We show more qualitative examples in Fig.~\ref{fig:qual_more}. In Fig.~\ref{fig:444polevaule}, there are several occurrences of Pole Vault activity, and our method can capture most of them. We show some failure examples in Figure.~\ref{fig:188highjump} and Fig.~\ref{fig:85diving}. In Fig.~\ref{fig:188highjump}, our model erroneously captures some activities as high jump. In those erroneous segments, we observe that the person tends to do a high jump activity but restrain in the end without completing the full action. The same goes for Fig.~\ref{fig:85diving}. Previous WTAL approaches~\cite{islam2020weakly, wtalc} have also shown similar issues as an inherent limitation of WTAL methods. Because of the weakly-supervised nature, we infer that some errors related to incomplete activities are inevitable. 

